%% file: main.tex
\documentclass[10pt,twocolumn,letterpaper]{article}

\usepackage[pagenumbers]{cvpr} 

\input{preamble}

\definecolor{cvprblue}{rgb}{0.21,0.49,0.74}
\usepackage[pagebackref,breaklinks,colorlinks,citecolor=cvprblue]{hyperref}

\def\paperID{00275} 
\def\confName{CVPR}
\def\confYear{2024}

\def\modelname{LocVLM\xspace}

\title{Learning to Localize Objects Improves Spatial Reasoning in Visual-LLMs}

\author{
Kanchana Ranasinghe$^{1,2}$ \quad
Satya Narayan Shukla$^1$ \quad
Omid Poursaeed$^1$ \\
Michael S. Ryoo$^2$ \quad \quad
Tsung-Yu Lin$^1$ 
\vspace{0.5em} \\ 
$^1$Meta \quad $^2$Stony Brook University 
\vspace{0.1em} \\
{\tt\small \{kranasinghe, mryoo\}@cs.stonybrook.edu, \{satyanshukla, opoursaeed, tsungyulin\}@meta.com}
}

\begin{document}
\maketitle

\input{sec/0_abstract}    
\input{sec/1_intro}
\input{sec/2_related}

\input{sec/3_method}
\input{sec/4_exp}

\input{sec/5_conclusion}

{
    \small
    \bibliographystyle{ieeenat_fullname}
    \bibliography{main}
}

\input{sec/X_suppl}

\end{document}

%% file: preamble.tex
\usepackage[dvipsnames]{xcolor}

\newcommand{\todo}[1]{{\color{red}#1}}

\usepackage[indent]{parskip}
\usepackage{multirow}

\usepackage{pifont}
\newcommand{\cmark}{\ding{51}}%
\newcommand{\xmark}{\ding{55}}%

\usepackage{indentfirst}

\usepackage{colortbl}
\definecolor{Gray}{gray}{0.90}

%% file: sec/0_abstract.tex
\begin{abstract}
\vspace{-0.5em}
Integration of Large Language Models (LLMs) into visual domain tasks, resulting in visual-LLMs (V-LLMs), has enabled exceptional performance in vision-language tasks, particularly for visual question answering (VQA). However, existing V-LLMs (e.g. BLIP-2, LLaVA) demonstrate weak spatial reasoning and localization awareness. Despite generating highly descriptive and elaborate textual answers, these models fail at simple tasks like distinguishing a left vs right location. 
In this work, we explore how image-space coordinate based instruction fine-tuning objectives could inject spatial awareness into V-LLMs. We discover optimal coordinate representations, data-efficient instruction fine-tuning objectives, and pseudo-data generation strategies that lead to improved spatial awareness in V-LLMs. Additionally, our resulting model improves VQA across image and video domains, reduces undesired hallucination, and generates better contextual object descriptions. 
Experiments across 5 vision-language tasks involving 14 different datasets establish the clear performance improvements achieved by our proposed framework. 
\vspace{-1.0em}
\end{abstract}

%% file: sec/1_intro.tex
\section{Introduction}
\label{sec:intro}

Holistic visual understanding requires learning beyond simply content of an image to encompass awareness on spatial locations of objects and their relations \cite{marr1982vision}. In the context of visual question answering (VQA), such spatial awareness allows better reasoning involving structural and contextual information contained within an image \cite{chen2023shikra}.

Since the introduction of powerful large-language models (LLMs) such as GPT-3 \cite{brown2020language}, Chat-GPT \cite{gpt4}, Vicuna \cite{vicuna2023}, and LLaMA~\citep{touvron2023llama,touvron2023llama2} that are capable of human style conversation, their visual counterparts such as BLIP-2 \cite{li2023blip}, LLaVA \cite{liu2023visual} have enabled novel tasks within the vision modality. However, despite their \cite{li2023blip,liu2023visual} highly generic visual understanding, these models exhibit poor language-based spatial reasoning \cite{chen2023shikra}. In fact, they fail at simple tasks such as distinguishing whether an object lies to the left or right of another object (see \cref{tbl:spatial_icl}). 

\input{figures/intro}

In the case of contrastive language image models (such as CLIP \cite{radford2021learning}, ALIGN \cite{Jia2021ScalingUV}), recent works explore how injecting explicit spatial awareness \cite{Zhang2023AssociatingSG,Luo2022SegCLIPPA,Mukhoti2022OpenVS,Ranasinghe2022PerceptualGI} can enable more holistic visual understanding. In fact, \cite{Ranasinghe2022PerceptualGI} shows how such improved spatial awareness benefits model robustness in adversarial domains. 
This raises the question of how generative language image models, particularly those connecting LLMs to visual encoders \cite{li2023blip,liu2023visual} can benefit from such spatial awareness specific training. We refer to models of this category that generate textual outputs given joint image-text inputs (e.g. \cite{li2023blip,liu2023visual}) as visual-LLMs (V-LLMs). 

In this work, we explore location specific instruction fine-tuning objectives that explicitly enforce V-LLMs to meaningfully process and generate textual image-space coordinates. We hypothesize that such training would lead to improved spatial awareness in these V-LLMs, therein improving performance on VQA tasks. To this end, we propose three instruction fine-tuning objectives that unify location representation with natural language. We also explore optimal representation forms for image-space locations and how pseudo-data generation can be leveraged for efficient scaling of our framework. We name our resulting model as \modelname.  

While the idea of adapting V-LLMs to perform localization related tasks (e.g. detection, segmentation) using V-LLMs has been explored in multiple recent works \cite{zhang2023gpt4roi,zhao2023bubogpt,zang2023contextual,peng2023kosmos,You2023FerretRA,wang2023visionllm,lai2023lisa}, these approaches depend on task specific architectural modifications or treat localization inputs / outputs differently from natural language. In contrast, our \modelname focuses on a unified framework treating location and language as a single modality of inputs with the goal of complementing performance in each task. We intuit that processing location represented in textual form would enforce the LLM to select appropriate image regions as opposed to relying on region level features provided by the architecture. At the same time, textual form location outputs promote spatial awareness at language level in a human interpretable manner, in contrast to using secondary heads or specialized tokens for location prediction.  
Concurrent work in \cite{chen2023shikra} also explores textual location representation with a generic V-LLM architecture similar to our work. Our proposed \modelname differs with focus on optimal location representation forms, data-efficient pseudo-labelling, and video domain operation. 

Our proposed framework exhibits improved spatial awareness in VQA style conversation demonstrated through experimentation on 14 datasets across 5 vision-language tasks: Spatial Reasoning, Image VQA, Video VQA, Object Hallucination, and Region Description. 
We summarize our key contributions as follows: 
\begin{itemize}[leftmargin=1em,noitemsep, topsep=0.0ex,itemsep=-1.0ex,partopsep=0ex,parsep=1ex]
    \item Inject textual spatial coordinate awareness into V-LLMs 
    \item Propose three novel localization based instruction fine-tuning objectives for V-LLMs 
    \item Discover optimal coordinate representation forms 
    \item Pseudo-Data generation for improved region description and scaling to video domain
\end{itemize} 

%% file: figures/intro.tex
\begin{figure}
\centering
\includegraphics[width=0.90\linewidth]{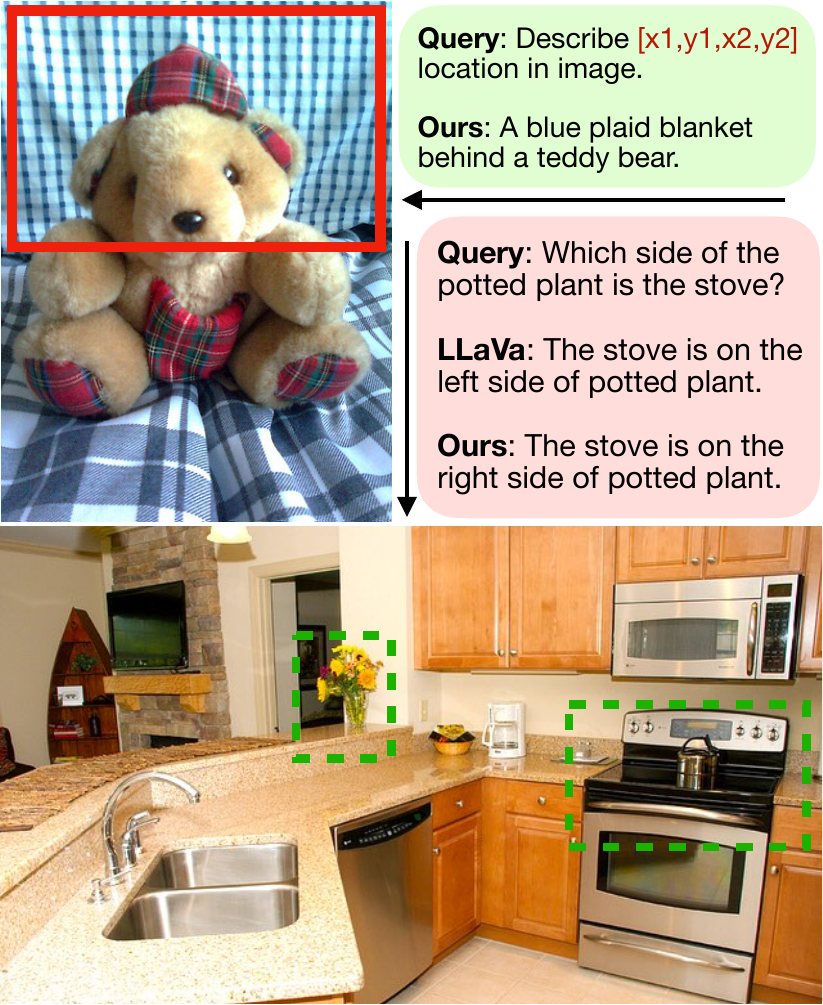}
\vspace{-0.5em}
\caption{We illustrate one unique ability of our model: contextual region description (top). Note the contextual information used in describing the selected region in each image. Explicitly teaching localization to Visual-LLMs also improves spatial awareness in VQA settings (bottom). Color boxes only for illustration purposes.}
\label{fig:intro}
\vspace{-1em}
\end{figure}

%% file: sec/2_related.tex
\section{Related Work}
\label{sec:related}

\noindent
\textbf{Localization in Contrastive Vision Language Models}:
Foundation vision language models (VLMs) such as CLIP \cite{radford2021learning} resulted in extensive exploration into language-tied localization in images both under dense (pixel / bounding-box) supervision \cite{gu2021vild, kamath2021mdetr, Li2022AdaptingCF, Li2021GroundedLP, Zeng2021MultiGrainedVL,Dou2022CoarsetoFineVP,ghiasi2022open, li2022language, Zhang2022GLIPv2UL,Ding2021DecouplingZS} and weak supervision \cite{xu2022groupvit, Xu2023LearningOS, Zhou2021ExtractFD, yao2022filip,cui2022democratizing, Zhang2023AssociatingSG,Luo2022SegCLIPPA,Ranasinghe2022PerceptualGI,Mukhoti2022OpenVS}. Recovering explicit localization information within model representations has enabled more robust operation for certain tasks \cite{Ranasinghe2022PerceptualGI}. While our work differs from this contrastive setting given our use of LLM based generative predictions, we similarly explore how explicit location information within the language modality can improve V-LLMs.  

\noindent
\textbf{Visual Large Language Models (V-LLMs)}:
The advent of powerful large language models (LLMs) such as GPT-3 \cite{brown2020language}, Chat-GPT \cite{gpt4}, and PaLM \cite{chowdhery2022palm}, as well as their open-source counterparts BLOOM~\citep{scao2022bloom}, Vicuna \cite{vicuna2023}, and LLaMA~\citep{touvron2023llama,touvron2023llama2}, has resulted in direct use of these LLMs for computer vision tasks \cite{Gupta2022VisualPC, Suris2023ViperGPTVI}. Alternate lines of work explore how LLMs can be connected to existing visual foundation models \cite{alayrac2022flamingo,anas_awadalla_2023_7733589,Ranasinghe2023LanguagebasedAC,li2023blip,liu2023visual,Maaz2023VideoChatGPT}, in particular to CLIP visual backbones \cite{radford2021learning}. While earlier models explored large-scale (millions to billions of samples) image-text training \cite{alayrac2022flamingo,anas_awadalla_2023_7733589}, later models \cite{li2023blip,liu2023visual,Maaz2023VideoChatGPT} scale down on data dependency. LLaVA \cite{liu2023visual} in particular scales down on pre-training data to under 1 million image-text pairs, and use instruction fine-tuning \cite{Wei2021FinetunedLM} to enable human-style conversation with visual awareness. This is extended to video domain in \cite{Maaz2023VideoChatGPT,ranasinghe2024understanding}. A shortcoming of these models is their lack of spatial awareness or location understanding in image space \cite{chen2023shikra,Gokhale2022BenchmarkingSR,Cho2022DALLEVALPT}.
Spatial reasoning limitations in generative VLMs are studied in \cite{Gokhale2022BenchmarkingSR,Cho2022DALLEVALPT}. Similar failures in captioning (and VQA) models are explored in \cite{Kamath2023WhatsW}. A solution in \cite{Hsu2023WhatsLC} proposes code-generation based reasoning. Our work tackles these same limitations but follows an alternate direction of spatial-aware instruction fine-tuning.
Another line of recent works \cite{zhang2023gpt4roi,zhao2023bubogpt,zang2023contextual,peng2023kosmos,You2023FerretRA,wang2023visionllm,lai2023lisa} tackle this by introducing architectural modifications to explicitly extract region level features that are injected to the LLM as special tokens. While introducing extra tokens and layers, this also separates the localization task from language. 
In contrast, we use a generic architectures with purely textual location information (i.e. image space coordinates as text).  
Concurrent work in \cite{chen2023shikra} explores this same idea, but we differ in 3 ways with, a) focus on optimal coordinate representation forms, b) data-efficient pseudo-labelling strategies, and c) video domain operation (see also \cref{tbl:related}). 

\input{figures/related}

\noindent
\textbf{Location Representations}:
Selecting regions within an image has a rich history in computer vision \cite{malik2001visual,uijlings2013selective} with greater focus on location outputs since the popularity of object detection \cite{girshick2015fastrcnn, redmon2016yolo, tian2019fcos, carion2020detr, chen2022diffusiondet, chen2021pix2seq, wang2023visionllm}. Early anchor-based methods regress locations from anchor centers \cite{girshick2015fastrcnn, redmon2016yolo}, followed by direct location regression from object-level features \cite{tian2019fcos, carion2020detr}. Recent works explore generative location predictions with diffusion processes \cite{chen2022diffusiondet} and sequence-generation \cite{chen2021pix2seq, wang2023visionllm}. Ours resembles the latter given our use of an LLM, next token prediction objective, and sequential generation of textual location representations. However, \cite{chen2021pix2seq, wang2023visionllm} utilize 1000 specialized location tokens (introduced to the LLM vocabulary) corresponding to 1000 bins uniformly spread across image space. While we explore similar binning strategies, in contrast we introduce no additional tokens, focus on purely textual representation of locations, and explore multiple textual location representation forms.

%% file: figures/related.tex
\begin{table}[t]
\centering
\small
\def\arraystretch{1.0}  
\setlength\tabcolsep{0.5em}  
\scalebox{0.95}{
\begin{tabular}{l|c|c|c|c}
\toprule
Method          & Kosmos \cite{peng2023kosmos} & Ferret \cite{You2023FerretRA} & Shikra \cite{chen2023shikra} & Ours \\ \midrule
Unified Arch.   & \xmark & \xmark & \cmark & \cmark  \\ 
Purely Textual  & \xmark & \xmark & \cmark & \cmark  \\ 
Pseudo Data     & \xmark & \xmark & \xmark & \cmark  \\ 
Video Domain    & \xmark & \xmark & \xmark & \cmark  \\ \bottomrule
\end{tabular}
}
\vspace{-0.5em}
\caption{Related Work Comparison: 
A unified architecture, purely textual inputs, pseudo data for scalable learning, and video domain operation distinguishes our work from these prior methods.}
\label{tbl:related}
\vspace{-1.0em}
\end{table}

%% file: sec/3_method.tex
\section{Method}
\label{sec:method}
Current V-LLMs \cite{liu2023visual,li2023blip} exhibit weak understanding of spatial locations within images \cite{chen2023shikra}. We explore and benchmark such shortcomings, and propose three novel instruction fine-tuning objectives aimed at overcoming these drawbacks of existing V-LLMs. We build these objectives based on spatial-coordinate based prompting and demonstrate how LLMs can directly both process and generate meaningful numerical coordinates in image-space after suitable training. In the rest of this section we describe our architecture and training framework, followed by coordinate processing \& generation, instruction fine-tuning objectives, pseudo-data generation, and video domain operation.   

\subsection{Architecture and Training}
\label{subsec:arch}
The focus of our work is to explore how spatial localization related training can improve a generic V-LLM such as LLaVA \cite{liu2023visual}. Therein, our architecture and training framework is inspired from \cite{liu2023visual}. We use a visual encoder, adapter layer, and LLM stacked sequentially (illustrated in \cref{fig:llava}), and follow a multi-stage training strategy similar to \cite{liu2023visual}. 

Consider an image $X \in \mathbb{R}^{H,W,C}$ where $H,W,C$ ($=3$) denote height, width, channels of image and a textual prompt $T$ composed of natural language (asking a question about the image). We define two variants of our model, \modelname-B and \modelname-L for better comparison with prior work. We first describe \modelname-B that processes images with $H=W=224$. Our visual encoder, ViT-L/14 from CLIP \cite{radford2021learning}, processes the image $X$ to produce a set of $256$ visual tokens in $\mathbb{R}^{1024}$, which are in turn projected to $\mathbb{R}^{4096}$ by an adapter layer (implemented as a linear layer). The LLaMA \cite{touvron2023llama} text tokenizer processes the textual prompt $T$ to produce textual tokens in $\mathbb{R}^{4096}$. The joint set of visual and textual tokens (of dimension $\mathbb{R}^{4096}$) are processed by a LLaMA \cite{touvron2023llama} LLM to produce the final set of textual tokens which are in turn untokenized to convert to natural language. The final natural language output is expected to be a suitable response to the input textual prompt, $T$. In variant \modelname-L, we use images sized $H=W=336$ resulting in 576 visual token, an adapter layer implemented as an MLP, and the LLM from LLaMA-2 \cite{touvron2023llama2}. All other design choices remain unchanged.   

We also highlight the BPE tokenization that is employed in our setup. This learned tokenization scheme may split a single word into sub-parts (that alone can appear meaningless to humans) and handles numerical text (including decimal point) as individual tokens (e.g. $12.34$ would be split into 5 separate tokens). 

In terms of training, we follow a two-stage strategy. Inspired by LLaVA \cite{liu2023visual}, we adopt an initial pre-training stage that only updates weights of the intermediate adapter layer to align the visual encoder outputs with LLM inputs. Next, we jointly instruction fine-tune the adapter layer and LLaMA LLM with our proposed objectives and template-based localization datasets (see \cref{subsec:exp_setup}). 
Video domain operation introduces an additional phase (see \cref{subsec:video}). 

\input{figures/llava}

\subsection{Coordinate Processing and Generation}
Humans contain the ability to reason about images using image-space coordinates. This is in contrast to existing V-LLMs that can describe the contents of an image elegantly, but lack spatial awareness regarding image contents. 
We hypothesize that injecting LLMs with additional spatial awareness, through coordinate based reasoning could improve their generic reasoning ability as well. To this end, we introduce our first goal of directly using textual coordinate based image locations in both natural language prompts and LLM generated outputs. For textual coordinates, we explore three different representation forms:
\begin{enumerate}[leftmargin=3em,noitemsep,topsep=-0.2ex,itemsep=-1.0ex,partopsep=0ex,parsep=1ex]
    \item Normalized Floating Point Values
    \item Integer Valued Binning (across image dimensions)
    \item Deviation from Image-Grid based Anchors 
\end{enumerate} 

\input{figures/ablate_coord}

\noindent
For image locations, we explore point based (e.g. center coordinates [cx, cy] of object) and bounding box based (e.g. top-left and bottom-right extreme coordinates of object region [x1, y1, x2, y2]) forms. We next discuss the three representations for coordinates used for either location. 

\noindent
\textbf{Normalized Floating Point Values} calculates absolute image coordinates and normalizes with image dimensions to a (0, 1) range. We use a 4 decimal point representation for these floating point values. While this representation is simple and generic, given the nature of BPE tokenization, each individual coordinate will be represented by up to 6 tokens. 

\noindent
\textbf{Integer Valued Binning} discretizes the absolute image coordinates to one of $n_b$ (=224, 336 for variant B \& L respectively) bins spread uniformly across the two image dimensions. Based on the binning parameter, $n_b$, each coordinate will be represented some number of tokens, in our case up to 3 (less than the floating point variant).

\noindent
\textbf{Deviation from Image-Grid based Anchors} is motivated from prior object detection works that estimate an initial anchor followed by deviation from that anchor center to estimate bounding box coordinates. We follow a similar setup, where one of $n_a$ anchors is predicted by the model, followed by deviation of coordinate from that anchor center. Our intuition is that, given the sequential next-token prediction setup of LLMs, such a two-stage strategy would lead to faster learning and more accurate coordinates. 

We refer to \cref{supp:coord} for further details on each variant. In \cref{tbl:ablate_coord}, we ablate each representation format on three different tasks (see \cref{subsec:ablation} for more details) of image VQA (GQA), region description (RD), and video VQA (A-QA). Our experiments indicate optimal performance for integer valued binning (IVB). In all following experimentation, we fix our coordinate representation to IVB. 

\subsection{Instruction Fine-Tuning Objectives}
\label{subsec:train_obj}
Given suitable coordinate representations, we now have a mechanism to directly prompt LLMs with image locations in textual form. Our second goal is to build training objectives using these image coordinates that directly inject spatial awareness into V-LLMs. We propose three instruction fine-tuning objectives for this purpose. 

Let us first revisit the visual instruction fine-tuning methodology in \cite{liu2023visual}. Building off the COCO dataset, they construct a VQA dataset containing conversation style question-answer pairs relevant to each COCO image. Question-answer pairs are generated using an LLM that is fed with the ground-truth bounding-box annotations for each image. Inspired by this setup, we build a similar spatial-VQA dataset using images and annotations of the COCO dataset, but instead of LLM prompting, we utilize hand-crafted templates and pseudo-captions (discussed in \cref{subsec:pseudo}) to generate conversations. 

We propose three types of question-answer pairs that relate to our three instruction fine-tuning objectives: Location Prediction (LocPred), Negative Prediction (NegPred), and Reverse-Location Prediction (RevLoc). 
See \cref{tbl:ift-obj} for examples.
Considering the LLM based final text generation in our architecture, we utilize next-token prediction loss to achieve each objective during our training. 

\noindent \textbf{Location Prediction}: 
Given an object category, we query the model to generate a point or bounding box localizing that object in the image. The object category and bounding box are derived from the COCO train set annotations. To avoid object mismatches (i.e. multiple object of same category), we first filter images containing only a single object of a given class. 

\noindent \textbf{Negative Prediction}: Using the same prompt templates as in \textit{LocPred} above, we query the model to generate a point or bounding box localizing a specified object in the image. However, in this case we select an object category not present in the image and accordingly provide a target text of ``no such object in image''. For each image, we utilize COCO bounding-box annotations to discover objects (belonging to COCO classes) that are not present in that image. 

\noindent \textbf{Reverse-Location Prediction}: We perform the reverse of \textit{LocPred} here. Given a point or bounding box in image space, we query the model to describe the object in that location. The bounding box and object category are derived from the COCO train set annotations. 

While introducing three novel train objectives aimed at injecting location information, we highlight that our proposed framework relies on training data (i.e. human annotations) identical to those used by LLaVA \cite{liu2023visual}. We do not use any additional ground-truth annotations for training. Next we explore how we could augment the generality of our framework while limiting to this same annotated data. 

\input{figures/ift_obj}

\subsection{Pseudo-Data Generation}
\label{subsec:pseudo}

We introduced three train objectives, each utilizing template based conversations as prompts and targets. However, our reliance on categories of COCO dataset limits the object vocabulary seen during training. Therein, we propose a pre-trained V-LLM based pseudo-data generation strategy. 
In fact, we utilize our model after stage one training as the V-LLM
leading to a form of self-training based learning. Given the abundance of only image-level annotated datasets (i.e. no bounding box ground-truth), we also explore how an object-detector generated pseudo-bounding boxes could augment our framework. 

\noindent \textbf{Self-Training}:
Given an image and bounding box annotations from the COCO train set, we prompt the V-LLM to caption each distinct object in the image. In order to prevent ambiguous object queries, we filter images to select only those containing at most one instance of a single category. We additionally prompt the V-LLM to describe the object using relational information (i.e. relative to other objects in image). This process provides us a dataset with object level bounding boxes and descriptive captions that are not limited to the COCO object categories (dataset details in \cref{supp:dataset}). In turn, we use this data generated by the V-LLM (our stage one model) to further improve performance of our framework.
We modify each of our three train objectives (in \cref{subsec:train_obj}) to utilize these image-specific pseudo-captions instead of the generic dataset level category labels.  

\noindent \textbf{Weak Supervision}: 
We explore how datasets containing no object level annotation (e.g. video classification / VQA datasets) could be leveraged to adapt our framework into domains beyond images. Therein, we utilize an off-the-shelf panoptic segmentation framework from SEEM \cite{Zou2023SegmentEE} to generate pseudo-bounding boxes for selected object categories within any image as well as exhaustive pixel level labels (enabling negative class identification). We leverage this setup to extend our introduced train objectives to the video domain as well. 


\subsection{Video Domain Operation}
\label{subsec:video}
Inspired by the simple modifications to LLaVA \cite{liu2023visual} in \cite{Maaz2023VideoChatGPT} enabling video domain operation, we follow a similar strategy of modifying our \modelname-B architecture to process videos while introducing no additional components. The visual backbone process multiple video frames individually (as images) and resulting tokens are averaged across spatial ($S$) and temporal ($T$) axes to obtain $S+T$ tokens. These are processed by the adapter layer and LLM to generate the textual outputs. Further details on our video architecture are discussed in \cref{supp:video}. 

In addition to our two training phases discussed in \cref{subsec:arch}, we introduce a third video instruction fine-tuning stage using a dataset we derive from ActivityNet \cite{Heilbron2015ActivityNetAL}. 
Following \cite{Maaz2023VideoChatGPT}, only the adapter layer is fine-tuned leaving all other parameters frozen. This resulting model is referred to as \modelname-Vid-B. 

We next introduce video variants of our three instruction fine-tuning objectives focused on static objects in videos. We utilize our proposed pseudo-labeling strategy to generate necessary video annotations and train both the adapter layer and LLM to obtain a second video model tagged \modelname-Vid-B+. Futher details on our video fine-tuning objectives are presented in \cref{supp:video}. 

%% file: figures/llava.tex
\begin{figure}[t]
\centering
\vspace{1em}
\includegraphics[width=0.42\textwidth]{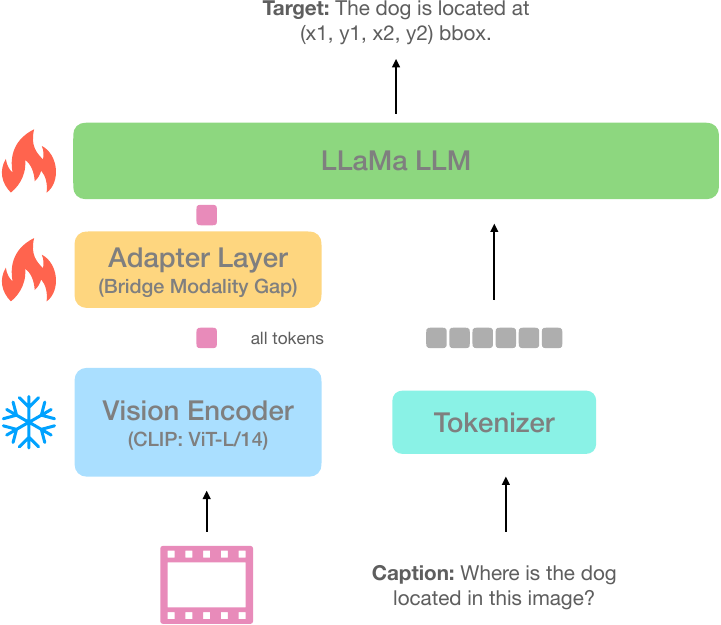}
\caption{
\textbf{Architecture:} We present the overall model architecture of our framework which is inspired from LLaVa \cite{liu2023visual}.
}
\label{fig:llava}
\vspace{-1.0em}
\end{figure}

%% file: figures/ablate_coord.tex
\begin{table}[t]
\centering
\small
\def\arraystretch{1.0}  
\setlength\tabcolsep{0.9em}  
\scalebox{0.97}{
\begin{tabular}{l|c|c|c}
\toprule
CR   & GQA (Acc)  & RD (METEOR)  & A-QA (Acc)\\ \midrule
NFP  & 46.1 & 19.6 & 37.1 \\ 
\rowcolor{Gray}
IVB  & 47.3 & 20.7 & 37.4 \\ 
DIGA & 47.0 & 20.8 & 37.3 \\ \bottomrule
\end{tabular}
}
\vspace{-1.0em}
\caption{Ablation on Coordinate Representation (CR) methods: we compare each of the three proposed CR variants, namely normalized floating point values (NFP), integer valued binning (IVB), and deviation from image-grid based anchors (DIGA).}
\label{tbl:ablate_coord}
\vspace{-1.0em}
\end{table}

%% file: figures/ift_obj.tex
\begin{table}[t]
\centering
\small
\def\arraystretch{1.0}  
\setlength\tabcolsep{0.9em}  
\begin{tabular}{l|c|c}
\toprule
Objective  & Prompt             & Target                \\ \midrule
LocPred    & Where is obj1?     & It’s at (x1,y1,x2,y2). \\
NegPred    & Where is obj2?     & There’s no obj2.       \\
RevLoc     & Describe (cx,cy)   & \textit{Detailed description}  \\ \bottomrule
\end{tabular}
\vspace{-1.0em}
\caption{We summarize our three distinct instruction fine-tuning objectives. Refer to appendix (\cref{supp:prompt}) for exact natural language prompts and targets used for training. For illustration, we use both point  and bounding-box based image locations here.}
\label{tbl:ift-obj}
\vspace{-1.0em}
\end{table}

%% file: sec/4_exp.tex
\section{Experiments}
\label{sec:exp}

In this section, we present experimental results to highlight existing weaknesses of SOTA V-LLMs and how our proposed framework addresses these issues. We also evaluate on standard VQA benchmarks across domains to showcase the better reasoning abilities of our model and highlight novel abilities of our framework. 

\subsection{Experimental Setup}
\label{subsec:exp_setup}
\textbf{Datasets}: 
We utilize the COCO dataset \cite{lin2014microsoft} and our model (post stage one training) to construct a localization related VQA dataset as outlined in \Cref{subsec:train_obj,subsec:pseudo}. We name this dataset \textit{Localize-Instruct-200K}. In detail, this contains LocPred and RevLoc question-answer pairs that use pseudo-captions instead of COCO categories as well as NegPred. We define a second video dataset, \textit{Localize-ActivityNet}, containing question-answer pairs constructed from Activity-Net pseudo-bounding boxes following \Cref{subsec:video}. 
Our models are primarily trained on our Localize-Instruct-200K dataset. Our stage one training uses CC3M dataset \cite{Changpinyo2021Conceptual1P}. 
Additionally, our Localize-ActivityNet dataset and ActivityNet dataset \cite{Heilbron2015ActivityNetAL} are used for video domain training. 

\noindent \textbf{Training}: We train our models on 8xA100 GPUs (each 80GB) following a two-phase training schedule. Our first phase trains on CC3M \cite{Changpinyo2021Conceptual1P} following the setup in \cite{liu2023visual}. 
The second phase uses our Localize-Instruct-200K dataset and trains for 10 epochs with a batch size of 64, ADAM-W optimizer with initial learning rate $2e-5$, 0.3 warm-up ratio, and cosine-decay learning rate schedule. 
Both the training phases we conduct use standard next-token-prediction loss used in LLM training. 

\noindent \textbf{Evaluation}: During evaluation, following standard protocol \cite{liu2023visual}, we iteratively generate next tokens, given visual and textual inputs. The LLM output is a distribution across the entire token vocabulary. The next token is selected through multinomial sampling of this output using a softmax temperature term of 0.2 during normalization.

\subsection{Spatial Reasoning: A Toy Experiment}
\label{subsec:spatial}
We investigate spatial reasoning abilities of two SOTA V-LLMs, LLaVA \cite{liu2023visual} and BLIP-2 \cite{li2023blip}, using a simple toy experiment. We create an evaluation dataset from COCO annotations containing images with distinct category object triplets (only one instance occurrence of each object category), where each object is entirely to the left or right half of the image and two objects are on opposite sides. 
The ground-truth bounding box annotation are utilized to automate this dataset creation procedure. 
This evaluation set, referred as \textit{COCO-Spatial-27K}, contains 26,716 image-question pairs (see \cref{supp:dataset} for details). 
We introduce two evaluation settings, direct VQA and in-context learning (ICL) VQA to understand spatial reasoning abilities of these models. In direct VQA, given an image we query the model whether an object lies above or below another object. In ICL VQA, before a similar final query, we provide two example question-answer pairs (involving the other two objects in the image) in the same format as our query. 
Refer to \cref{supp:toy} for further details on task. We perform the same for objects in top vs bottom halves of images. 

These results are presented in \Cref{tbl:spatial_icl}. Our results indicate near random performance for existing V-LLMs. 
For the case of LLaVA, we perform keyword (\textit{left} and \textit{right}) frequency analysis on its instruction tuning dataset (LLaVA-Instruct-80K dataset) to verify the presence of terms \textit{left} and \textit{right} in its training corpus. These keywords are present in 0.37\% and 1.13\% of its conversations respectively (see \cref{supp:llava_dataset} for more) indicating presence of these concepts in the image-text training corpus. In contrast to these methods, our proposed framework notably improves performance over both BLIP-2 \cite{li2023blip} and the LLaVA baseline \cite{liu2023visual}.

\subsection{Image VQA}
Image VQA involves correctly answering natural language questions regarding content within an image. We evaluate our model for Image VQA on two standard datasets, GQA and VQAv2. The GQA dataset focuses on questions requiring compositional reasoning, particularly involving surrounding information of objects within an image. We evaluate on its test-dev split containing 12,578 image-question pairs. The VQAv2 dataset contains open-ended questions about each image that require an understanding of vision, language and commonsense knowledge to answer. We use its validation split containing 214,354 image-question pairs for our evaluation. 
For each dataset, we follow standard V-LLM evaluation protocol following \cite{li2023blip,Maaz2023VideoChatGPT} and report top-1 accuracy metric. Our results in \cref{tbl:res_im_vqa} indicate clear improvements for \modelname over our baseline and prior work, establishing the usefulness of our proposed framework. The closest to our work, Shikra \cite{chen2023shikra} achieves performance competitive to our \modelname-B, but unlike ours they use VQA datasets (containing similar domain question-answer pairs) during training. 

\input{figures/spatial_icl}
\input{figures/eval_im_vqa}
\input{figures/eval_vid_vqa}

\subsection{Video VQA}
Our model is also applicable to video tasks following our video domain adaptation described in \cref{subsec:video}. We simply adopt the additional video instruction fine-tuning phase from \cite{Maaz2023VideoChatGPT} on the ActivityNet dataset after our initial two phases of training to obtain \modelname-Vid-B. This third phase involves fine-tuning only the adapter layer of our model. We also explore video variants of our IFT objectives that train both adapter layer and LLM. The resulting model is termed \modelname-Vid-B+. 

Video VQA focuses on correctly answering questions regarding a given video that require spatio-temporal awareness to answer. 
We evaluate our video-adapted model on the task of zero-shot video VQA on four benchmark datasets, ActivityNet-QA, MSRVTT-QA, MSVD-QA, and TGIF-QA. We evaluate on the validation splits of these four datasets. ActivityNet-QA videos cover a wide range of complex human activities relevant to daily living with its question-answer pairs focusing on long-term spatio-temporal reasoning. MSRVTT-QA builds off the MSRVTT dataset that contains web videos covering a comprehensive range of categories and diverse visual content. MSVD-QA is a similar dataset building off the MSVD dataset. TGIF-QA contains question-answer pairs from an dataset constructed of animated GIFs. For each dataset, we report the accuracy metric following evaluation protocol in \cite{Maaz2023VideoChatGPT}.
Our results on these four datasets reported in \cref{tbl:res_vid_vqa} demonstrate state-of-the-art performance of our proposed \modelname-Vid-B, with consistent improvements over the baseline from \cite{Maaz2023VideoChatGPT}. Here we use the \modelname-Vid-B variant for fairer comparison with the baseline from \cite{Maaz2023VideoChatGPT}. We attribute the performance gains exhibited by our model to its stronger spatial awareness (see \cref{subsec:spatial}). Particularly in the case of video understanding, awareness of content at spatial level of each frame is significant to understand object motions and interactions \cite{Ryoo2006RecognitionOC,Aggarwal2011HumanAA}. We also report more results involving additional model variants in \cref{tbl:vid_more}.

\subsection{Object Hallucination}
Current state-of-the-art V-LLMs suffer from object hallucination, generating image descriptions inconsistent with the image content \cite{li2023evaluating}. For example, a V-LLM would respond to ``Where is the cat in this image?" with ``The cat is on the table" when in reality there is no cat in the image. We evaluate the extent of hallucination in V-LLMs using three datasets we introduce (details in \cref{supp:dataset}) and the POPE dataset \cite{li2023evaluating}. Our three datasets, Hal-COCO, Hal-ADE, and Hal-Act build off COCO, ADE-20K, and ActivityNet datasets respectively. The first two involve images and the latter videeos. These datasets contain `Is there \textit{obj} in image / video?'' type questions per sample, for two objects present and not present in the image / video. Hal-ADE object categories contain \textit{no overlap} with COCO classes allowing evaluation on novel object categories unseen during our instruction fine-tuning. Results reported in \cref{tbl:res_hallucinate} show clear improvements of \modelname-B over baselines. We also evaluate \modelname-B on the POPE benchmark \cite{li2023evaluating} that builds off the COCO dataset object annotations and report results in \cref{tbl:pope}. Our \modelname showcases similar performance improvements on this dataset. 

\input{figures/eval_vid_vqa_more}
\input{figures/eval_hallucinate}
\input{figures/eval_pope}
\input{figures/eval_revloc}

\subsection{Region Description}
\label{subsec:revloc}
A unique characteristic of our model (in contrast to V-LLMs like LLaVA \cite{liu2023visual} \& BLIP-2 \cite{li2023blip}) is its ability to reason with prompts involving coordinate based image space locations without any input modifications. Given a point or bounding box location, we prompt our model to generate an output describing that location. We refer to this unique ability of our model as \textit{region description} (RD). We evaluate this RD capability of our model by generating object level descriptions focused on contextual information (e.g. surrounding of that object in the image). Following evaluation protocol in \cite{peng2023kosmos} for region description, we extend their evaluation to three standard referring localization datasets from RefCOCO \cite{kazemzadeh2014referitgame} and report these results in \cref{tbl:res_revloc}. We select the METEOR score as the evaluation metric to account for variations in word choice in generated answers which may be acceptable in various cases (e.g. different sentence structure leading to alternate word ordering). Our results indicate clear improvements over the LLaVA baseline \cite{liu2023visual} as well as prior state-of-the-art. We attribute these improvements to our pseudo-data based training.

\subsection{Ablations}
\label{subsec:ablation}
Next we conduct ablative studies on separate components of our proposed setup: IFT objectives, location type, and pseudo-data. We follow the same training strategy as described in \cref{subsec:exp_setup} and present these results in \cref{tbl:ablate_rest}. \modelname-B is used for all these experiments. The significance of each IFT objective is verified in \cref{tbl:ablate_rest} (top) with consistent performance improvements across tasks. 
The generality of our approach to differing location type (i.e. points vs bounding boxes) and usefulness of pseudo-data is visible in \cref{tbl:ablate_rest} (bottom). In particular, we highlight the notable performance improvement for RD task gained from using pseudo-data. 
We also conduct ablations for our video domain training setup and report these results in \cref{tbl:ablate_vid}. The \modelname-Vid-B+ variant is used in these experiments. Our results showcase the usefulness of proposed IFT objectives for video domain learning as well. 

\input{figures/ablate_all}
\input{figures/ablate_vid}

%% file: figures/spatial_icl.tex
\begin{table}[t]
\centering
\small
\def\arraystretch{1.0}  
\setlength\tabcolsep{0.6em}  

\scalebox{0.88}{
\begin{tabular}{l|c|ccc|ccc}
\toprule
Method                      & ICL    & All   & Left  & Right & All  & Above & Below \\ \midrule
BLIP-2 \cite{li2023blip}    & \xmark & 45.5  & 86.1  & 4.74  & 49.2 & 50.4  & 48.6  \\
LLava  \cite{liu2023visual} & \xmark & 55.1  & 84.5  & 36.5  & 58.9 & 57.8  & 59.3  \\ \rowcolor{Gray}
Ours                        & \xmark & 69.5  & 79.7  & 59.2  & 65.4 & 64.2  & 65.9  \\ \midrule
BLIP-2 \cite{li2023blip}    & \cmark & 14.7  & 17.8  & 11.6  & 15.8 & 16.5  & 15.2  \\ 
LLaVa \cite{liu2023visual}  & \cmark & 55.1  & 84.7  & 36.4  & 58.2 & 57.7  & 58.5  \\ \rowcolor{Gray}
Ours                        & \cmark & 76.5  & 90.4  & 61.5  & 74.1 & 73.5  & 74.4  \\ \bottomrule
\end{tabular}
}

\vspace{-0.5em}
\caption{\textbf{Spatial Reasoning}: We report accuracy (\%) on a spatial localization dataset derived from COCO annotations to highlight weak spatial awareness of existing V-LLMs. We query these models to answer whether one object is to the left or right / above or below of another object. The SOTA V-LLMs evaluated exhibit close to random performance. Our proposed setup outperforms existing methods. Ours refers to \modelname-B variant.}
\label{tbl:spatial_icl}
\vspace{-1.0em}
\end{table}

%% file: figures/eval_im_vqa.tex
\begin{table}[t]
\centering
\small
\def\arraystretch{1.0}  
\setlength\tabcolsep{0.6em}  
\scalebox{0.85}{
\begin{tabular}{l|c|c|c|c|c|c}
\toprule
Method & LLM & VS              & Zero-Shot     & GQA  & VQA-V & VQA-T \\ \midrule
SR \cite{Banerjee2021WeaklySR} & -  &  -  & \xmark & 62.1 & 72.9 & -    \\
Shikra \cite{chen2023shikra}   & 7B & 224 & \xmark & -    & 75.3 & 77.4 \\
LLaVA-v1.5                     & 7B & 336 & \xmark & 62.0 & 78.1 & 78.4 \\ \rowcolor{Gray}
\modelname-L                    & 7B & 336 & \xmark & \textbf{63.5} & \textbf{78.2} & \textbf{78.6} \\ \midrule
LLaVA-v1                       & 7B & 224 & \cmark & 44.7 & 49.8 & 49.3 \\ \rowcolor{Gray}
\modelname-B                    & 7B & 224 & \cmark & \textbf{47.3} & \textbf{50.3} & \textbf{50.8} \\ \midrule
Viper-GPT                      & 175B & - & \cmark & 48.1 & -    & -    \\
BLIP-2                         & 11B& - & \cmark & 44.7 & 54.3 & 53.9 \\
LLaVA-v1.5                     & 7B & 336 & \cmark & 48.7 & 55.7 & 55.3 \\ \rowcolor{Gray}
\modelname-L                    & 7B & 336 & \cmark & \textbf{50.2} & \textbf{55.9} & \textbf{56.2} \\ \bottomrule
\end{tabular}}
\vspace{-1.0em}
\caption{\textbf{Image VQA Results}: We report accuracy (\%) on the test-dev split of GQA dataset (GQA) and the validation / test splits of VQAv2 dataset (VQA-V / VQA-T). Our proposed \modelname improves over prior works achieving state-of-the-art performance.}
\label{tbl:res_im_vqa}
\vspace{-1em}
\end{table}

%% file: figures/eval_vid_vqa.tex
\begin{table*}[t]
\centering
\small
\def\arraystretch{1.1}  
\setlength\tabcolsep{1.3em}  
\scalebox{0.95}{
\begin{tabular}{l|c|c|c|c|c}
\toprule
Method          & Zero-Shot & ActivityNet-QA & MSRVTT-QA & MSVD-QA  & TGIF-QA \\ \midrule
JustAsk \cite{yang2021justask}        & \xmark    & 38.9            & 41.8      & 47.5     &  -      \\
FrozenBiLM \cite{yang2022zero}      & \xmark    & 43.2            & 47.0      & 54.8     &  -      \\
VideoCoCa \cite{Yan2022VideoCoCaVM}       & \xmark    & 56.1            & 46.3      & 56.9     &  -      \\ \midrule
Flamingo \cite{alayrac2022flamingo}        & \cmark    &   -             & 17.4      & 35.6     &  -      \\ 
BLIP-2 \cite{li2023blip}          & \cmark    &   -             & 17.4      & 34.4     &  -      \\ 
InstructBLIP \cite{dai2023instructblip}    & \cmark    &   -             & 25.6      & 44.3     &  -      \\
FrozenBiLM \cite{yang2022zero}      & \cmark    & 24.7            & 16.8      & 32.2     & 41.0    \\
Video Chat \cite{2023videochat}     & \cmark    & 26.5            & 45.0      & 56.3     & 34.4    \\
LLaMA Adapter \cite{Zhang2023LLaMAAdapterEF}   & \cmark    & 34.2            & 43.8      & 54.9     & -       \\
Video LLaMA \cite{Zhang2023VideoLLaMAAI}     & \cmark    & 12.4            & 29.6      & 51.6     & -       \\
Video-ChatGPT \cite{Maaz2023VideoChatGPT}   & \cmark    & 35.2            & 49.3      & 64.9     & 51.4    \\ \rowcolor{Gray}
\modelname-Vid-B  & \cmark    & \textbf{37.4}   & \textbf{51.2} & \textbf{66.1} & \textbf{51.8}     \\ \bottomrule
\end{tabular}
}
\vspace{-0.8em}
\caption{\textbf{Video VQA Results}: Our proposed \modelname-Vid-B improves over Video-ChatGPT \cite{Maaz2023VideoChatGPT} and achieves state-of-the-art results (Top-1 Accuracy \%) across four different video VQA benchmarks. Note the zero-shot setting of all these evaluations.} 
\label{tbl:res_vid_vqa}
\vspace{-1em}
\end{table*}

%% file: figures/eval_vid_vqa_more.tex
\begin{table}[t]
\centering
\small
\def\arraystretch{1.0}  
\setlength\tabcolsep{0.9em}  
\begin{tabular}{l|c|c|c}
\toprule
Method           & VLT  & Frames & Acc (\%) \\ \midrule
LLaVa (v1) \cite{liu2023visual}       & \xmark & 1   & 28.7 \\
LLaVa (v1.5) \cite{liu2023visual}     & \xmark & 1   & 31.5 \\ \rowcolor{Gray}
\modelname-B     & \xmark & 1   & 29.2 \\ \rowcolor{Gray}
\modelname-L     & \xmark & 1   & \textbf{32.1} \\ \midrule
Video-ChatGPT \cite{Maaz2023VideoChatGPT}   & \cmark & 100 & 35.2 \\ \rowcolor{Gray}
\modelname-Vid-B & \cmark & 100 & 37.4 \\ \rowcolor{Gray}
\modelname-Vid-B+& \cmark & 8   & \textbf{38.2} \\ \bottomrule
\end{tabular}
\vspace{-1.0em}
\caption{\textbf{Video VQA}: We report more results (Top-1 Accuracy) for ActivityNet-QA dataset including multiple baseline and \modelname variants. Our proposed models exhibit top performance. VLT denotes video level training. More details in \cref{supp:video}.}
\label{tbl:vid_more}
\vspace{-1em}
\end{table}

%% file: figures/eval_hallucinate.tex
\begin{table}[t]
\centering
\small
\def\arraystretch{1.0}  
\setlength\tabcolsep{0.9em}  
\scalebox{0.95}{
\begin{tabular}{l|c|c|c}
\toprule
Method        & Hal-COCO  & Hal-ADE & Hal-Act\\ \midrule
Shikra \cite{chen2023shikra}       & 86.2 & 58.7 & -   \\ 
LLaVa \cite{liu2023visual}        & 61.9 & 53.8 & - \\ \rowcolor{Gray}
\modelname-B  & \textbf{88.3} & \textbf{75.2} & - \\ \midrule
Video-ChatGPT \cite{Maaz2023VideoChatGPT}   & - & - & 50.6 \\ \rowcolor{Gray}
\modelname-Vid-B & - & - & 68.7 \\ \rowcolor{Gray} 
\modelname-Vid-B+& - & - & \textbf{72.4} \\ \bottomrule
\end{tabular}}
\vspace{-1.0em}
\caption{\textbf{Hallucination Evaluation}: We report top-1 accuracy (\%) for object presence type questions and showcase reduced object hallucination in our proposed framework.}
\label{tbl:res_hallucinate}
\vspace{-1em}
\end{table}


%% file: figures/eval_pope.tex
\begin{table}[t]

\newcolumntype{a}{>{\columncolor{Gray}}c}

\centering
\small
\def\arraystretch{1.0}  
\setlength\tabcolsep{0.6em}  
\scalebox{0.93}{
\begin{tabular}{l|l|ccca}
\toprule
Datasets & Metrics      & BLIP-2 & Shikra & LLaVA & Ours \\ \midrule 
\multirow{5}{*}{Random}
& Accuracy ($\uparrow$) & 88.6 & 86.9 & 50.4 & 87.9 \\
& Precision ($\uparrow$)& 84.1 & 94.4 & 50.2 & 83.6 \\
& Recall ($\uparrow$)   & 95.1 & 79.3 & 99.1 & 93.9 \\
& F1 Score ($\uparrow$) & \textbf{89.3} & 86.2 & 66.6 & 88.5 \\
& Yes                   & 56.6 & 43.3 & 98.8 & 56.2 \\ \midrule \multirow{5}{*}{Popular}
&Accuracy ($\uparrow$)  & 82.8 & 84.0 & 49.9 & 86.0 \\
&Precision ($\uparrow$) & 76.3 & 87.6 & 49.9 & 79.7 \\
&Recall ($\uparrow$)    & 95.1 & 79.2 & 99.3 & 93.9 \\
&F1 Score ($\uparrow$)  & 84.7 & 83.2 & 66.4 & \textbf{86.3} \\
&Yes                    & 62.4 & 45.2 & 99.4 & 58.9 \\ \midrule \multirow{5}{*}{Adversarial}
&Accuracy ($\uparrow$)  & 72.1 & 83.1 & 49.7 & 78.8 \\
&Precision ($\uparrow$) & 65.1 & 85.6 & 49.9 & 76.6 \\
&Recall ($\uparrow$)    & 95.1 & 79.6 & 99.1 & 93.7 \\
&F1 Score ($\uparrow$)  & 77.3 & 82.5 & 66.3 & \textbf{84.3} \\
&Yes                    & 73.0 & 46.5 & 99.4 & 61.7 \\ \bottomrule
\end{tabular}
}
\vspace{-1.0em}
\caption{\textbf{More object hallucination:} Results on POPE evaluation benchmark ~\citep{li2023evaluating} indicate strong performance of our model.}
\label{tbl:pope}
\end{table}

%% file: figures/eval_revloc.tex
\begin{table}[t]
\centering
\small
\def\arraystretch{1.0}  
\setlength\tabcolsep{0.58em}  
\scalebox{0.92}{
\begin{tabular}{l|c|c|c|cc}
\toprule
\multirow{2}{*}{Method} & \multirow{2}{*}{ZS} & \multirow{2}{*}{RefCOCO} & \multirow{2}{*}{RefCOCO+} & \multicolumn{2}{c}{RefCOCOg} \\
                              &        &   &   & Val  & Test \\ \midrule
SLR \cite{Yu2016AJS}          & \xmark & -    & -    &  -   & 15.4 \\
SLR + Rerank \cite{Yu2016AJS} & \xmark & -    & -    &  -   & 15.9 \\
Kosmos-2 \cite{peng2023kosmos}& \xmark & 8.67 & 8.82 & 14.3 & 14.1 \\
Shikra \cite{chen2023shikra}  & \xmark & 10.4 & 11.1 & 19.7 & 19.5 \\ 
LLaVa \cite{liu2023visual}    & \xmark & 8.43 & 8.73 & 13.5 & 13.5 \\ \rowcolor{Gray}
\modelname-B                  & \xmark & \textbf{14.6} & \textbf{15.2} & \textbf{26.0} & \textbf{26.2} \\ \midrule
Kosmos-2 \cite{peng2023kosmos}& \cmark & 6.34 & 8.25 & 12.4 & 12.2 \\
LLava \cite{liu2023visual}    & \cmark & 4.23 & 7.26 & 10.6 & 10.3 \\ \rowcolor{Gray}
\modelname-B                  & \cmark & \textbf{11.0} & \textbf{11.1} & \textbf{20.6} & \textbf{20.7}  \\ \bottomrule
\end{tabular}   
}
\vspace{-1.0em}
\caption{\textbf{Region Description:} We report METEOR scores for RD task \cite{peng2023kosmos}. Test-B split is used for RefCOCO \& RefCOCO+ datasets. Our method outperforms all prior work.} 
\label{tbl:res_revloc}
\vspace{-1.0em}
\end{table}

%% file: figures/ablate_all.tex
\begin{table}[t]
\begin{minipage}{1.0\linewidth}
\centering
\small
\def\arraystretch{1.0}  
\setlength\tabcolsep{0.7em}  
\scalebox{0.95}{
\begin{tabular}{c|c|c|c|c|c}
\toprule
LocPred & NegPred & RevLoc  & GQA & RD & A-QA \\ \midrule
\xmark  & \xmark  & \xmark  & 44.7 & 10.3 & 35.2 \\ 
\cmark  & \xmark  & \xmark  & 45.2 & 12.2 & 35.8 \\ 
\cmark  & \cmark  & \xmark  & 46.9 & 12.5 & 37.2 \\ \rowcolor{Gray}
\cmark  & \cmark  & \cmark  & 47.3 & 20.7 & 37.4 \\ \bottomrule
\end{tabular}}
\vspace{0.5em}
\end{minipage}
\begin{minipage}{1.0\linewidth}
\centering
\small
\def\arraystretch{1.0}  
\setlength\tabcolsep{1.1em}  
\scalebox{0.96}{
\begin{tabular}{l|c|c|c|c}
\toprule
Location Type    & PD     & GQA  & RD   & A-QA \\ \midrule
Point & \cmark & 47.3 & 20.6 & 37.4 \\ \rowcolor{Gray}
Bounding Box  & \cmark & 47.3 & 20.7 & 37.4 \\ 
Bounding Box  & \xmark & 46.5 & 11.6 & 37.1 \\ \bottomrule
\end{tabular}}
\vspace{-1.0em}
\end{minipage}
\caption{\textbf{Ablations:} We report top-1 accuracy (\%) on GQA and ActivityNet-QA (A-QA) datasets and METEOR scores for RD task on RefCOCOg test split. (top) We ablate proposed instruction fine-tuning objectives to verify usefulness of each objective. (bottom) We first ablate point based and bounding box based location forms to showcase minimal difference across them. We next ablate use of object description pseudo-data (PD). We highlight the improvements due to pseudo-data, especially on the RD task.}
\label{tbl:ablate_rest}
\vspace{-0.5em}
\end{table}

%% file: figures/ablate_vid.tex
\begin{table}[t]
\centering
\small
\def\arraystretch{1.0}  
\setlength\tabcolsep{1.0em}  
\scalebox{0.95}{
\begin{tabular}{c|c|c|c}
\toprule
LocPred & NegPred & RevLoc  & A-QA  \\ \midrule
\xmark  & \xmark  & \xmark  & 37.4  \\ 
\cmark  & \xmark  & \xmark  & 37.6  \\ 
\cmark  & \cmark  & \xmark  & 38.2  \\ \rowcolor{Gray}
\cmark  & \cmark  & \cmark  & 38.2  \\ \bottomrule
\end{tabular}}
\vspace{-1.0em}
\caption{\textbf{Video Ablation:} We report top-1 accuracy (\%) on ActivityNet-QA (A-QA) dataset. Results indicate the generality of proposed IFT objectives for video domain training as well.}
\label{tbl:ablate_vid}
\vspace{-1.0em}
\end{table}

%% file: sec/5_conclusion.tex
\section{Conclusion}
\label{sec:conclusion}

We introduce a simple framework that equips visual-LLMs (V-LLMs) with greater spatial understanding, termed \modelname. We leverage the idea of encoding image coordinates within language to propose three instruction fine-tuning (IFT) objectives. This training process endows V-LLMs with the ability to reason about spatial composition of images using image space coordinates within text. A data efficient training pipeline utilizing pseudo-data allows our approach to achieve state-of-the-art results in Image VQA, Video VQA, and Region Description while improving spatial awareness and reducing object hallucination. 


%% file: sec/X_suppl.tex
\clearpage
\setcounter{page}{1}
\maketitlesupplementary

\appendix

\section{Coordinate Representation Details}
\label{supp:coord}

We describe our three coordinate representation variants in detail, first focused on bounding-box location format . Consider an image of dimensions (512, 512) containing a cat. Let ($10, 120, 30, 145$) define the minimal bounding box enclosing the cat in image space ordered as (x1,y1,x2,y2) where (x1,y1) would describe the top left corner and (x2,y2) would describe the bottom right corner of that bounding box. We will use this example in following explanations. 

\noindent
\textbf{Normalized Floating Point Values} would normalize these coordinates using image dimensions to a (0,1) range and directly use  normalized values rounded to 4 decimal places. In the given example, the location of the cat would be described ($0.0195, 0.2344, 0.0586, 0.2832$) which is equal to ($10/512,120/512,30/512,145/512$) after appropriate rounding. 

\noindent 
\textbf{Integer Valued Binning} considers $n_b$ fixed bins across the image that are described by integers $0$ to $n_b$. In our case, for the \modelname-B version we fix $n_b$ to 224 and for \modelname-L version we fix $n_b$ to 336. The original bounding-box coordinates are mapped to the range $(0, n_b)$ inspired by prior work \cite{chen2021pix2seq, wang2023visionllm} using similar binning strategies. In the case of our examples, the location of the cat would be described ($4, 52, 13, 63$) for $n_b = 224$ which can be easily calculated by remapping the coordinate range as $(n_b \cdot 10/512,n_b \cdot 120/512,n_b \cdot 30/512,n_b \cdot 145/512$) with integer rounding. 

\noindent
\textbf{Deviation from Image-Grid based Anchors} defines a grid of anchors in image space, selects the anchor closest to the object center, and measures each bounding box coordinate as a deviation from that anchor center. In our case, we set $n_a = 16^2$ for \modelname-B and and $n_a = 24^2$ for \modelname-L (motivated by the visual encoder transformer grid size). In both cases, each anchor covers a $14 \times 14$ pixel patch. We describe the anchors using $(p,q)$ for $p,q = 0, 1, ... 13$. 
For our example, the bounding box fits the anchor ($0,4$) and we represent the bounding box as ($0, 4, 3, 11, 6, 0$) where the latter four values correspond to pixel deviations from the selected anchor center located at ($7,63$) in ($224 \times 224$) image space. 

We also utilize the alternate location form of point values, i.e. (cx, cy) for object center coordinates in image space. Coordinate representations are utilized in the same manner. Instead of four coordinates, we only use two that correspond to the object center. For our given example, the center of the cat would be ($20, 132.5$) which would be represented similar to the bounding box case.

\section{Training Prompt Details}
\label{supp:prompt}
We introduce three instruction fine-tuning objectives that utilize specific hand-crafted templates to generate the target prompts used during training. We discuss in detail, these three objectives presented in \cref{tbl:ift-obj} (main paper): LocPred, NegPred, and RevLoc. 

For the first two cases, we use a set of 5 templates, one of which is randomly selected for each sample during training. 
{\small
\begin{enumerate}[leftmargin=3em,noitemsep,topsep=-0.2ex,itemsep=-1.0ex,partopsep=0ex,parsep=1ex]
    \item \texttt{Where is the object described \{category\} located in image in terms of \{repr\}?}
    \item \texttt{What is the location of object described \{category\} in terms of \{repr\}?}
    \item \texttt{Localize the object described \{category\} in terms of \{repr\}?}
    \item \texttt{Provide a \{repr\} for the the object described \{category\}?}
    \item \texttt{Generate a \{repr\} for the the object described \{category\}?}
\end{enumerate} 
}
\noindent
The placeholder \texttt{\{category\}} is replaced with the relevant ground-truth annotation of each particular object. In the case of COCO dataset, these correspond to one of the 80 COCO categories. For Localize-Instruct-200K (our constructed pseudo-caption dataset), the object pseudo-description is used in place of \texttt{\{category\}}. The \texttt{\{repr\}} can be one of rep\_bbox = (x1,y1,x2,y2) bbox or rep\_point = (cx,cy) point.

For LocPred, the target is of form \texttt{``It is located at \{loc\}''} while for NegPred, the target is \texttt{``There is no such object in the image''}. The same five identical prompts are randomly assigned to each objective to ensure no input patterns allow distinguishing between the two targets.   

For the case of RevLoc, we similarly sample one prompt from the following set of 3 templates:
{\small
\begin{enumerate}[leftmargin=3em,noitemsep,topsep=-0.2ex,itemsep=-1.0ex,partopsep=0ex,parsep=1ex]
    \item \texttt{Describe the object located at \{loc\}?}
    \item \texttt{Provide a caption for object at \{loc\}?}
    \item \texttt{What is at location \{loc\} in image?}
\end{enumerate} 
}
The target is of form \texttt{``There is a \{category\}.''} where category can either be class label or a pseudo-description of that location.

\section{Dataset Details}
\label{supp:dataset}
In our work, we first perform blurring of human faces across all our data to preserve privacy in resulting models. These modifications are applied to all our datasets before performing any model training. 

As described in \cref{subsec:pseudo} (main paper), we explore pseudo-data generation to construct two new datasets, one for object level captions in images and the other for video object labels. We name them first PRefCOCO-100K, and utilize it to construct our Localize-Instruct-200K dataset used for our image level instruction fine-tuning (IFT) objectives. We name the second Pseudo-ActNet and utilize it in our video level IFT objectives. 

PRefCOCO-100K uses 95899 images from the COCO dataset and uses an image VQA model (LLaVa \cite{liu2023visual}) to generate object level descriptions using the COCO object annotations. We first filter images to select those containing unique instances of objects (e.g. only one dog in the image as opposed to multiple dogs). This results in the 95899 images. Next, we ask the VQA model to generate a suitable caption that describes the object category using both its characteristics and relations to surrounding. In detail, we use the exact prompt \texttt{``Describe the \{category\} in this image using one short sentence, referring to its visual features and spatial position relative to other objects in image.''} where \texttt{category} is the ground-truth object label. These obtained object-level captions are used to create question-answer (QA) pairs for the images, resulting in 402,686 such QA pairs. 

Following the prompting mechanisms for LocPred and RevLoc described in \cref{supp:prompt}, we generate image-conversation pairs from PRefCOCO-100K, resulting in a human-conversation style dataset we use for training. We refer to this dataset as Localize-Instruct-200K. This contains twice as many image-conversation pairs as the original, given repeated images for both LocPred and RevLoc objectives. This is the main dataset used for our image level training. 

For our video domain IFT objective based training, we only use category level labels and leave caption level training as a future direction. We construct Pseudo-ActNet dataset that contains generated bounding-box annotations for all objects belonging to COCO panoptic segmentation dataset \cite{lin2014microsoft} categories. Eight uniformly sampled frames are processes per video for annotation. We utilize the pre-trained SEEM \cite{Zou2023SegmentEE} model (motivated by \cite{li2023evaluating}) to generate pixel-level panoptic segmentation outputs for each selected frame and convert these segmentations to bounding boxes (panoptic also contains instance level distinction allowing straightforward bounding box extraction). The panoptic outputs (label for each pixel) also allows to obtain an exhaustive list of all COCO dataset categories present in each video - this is necessary to find suitable negative categories for our NegPred objective. Therein, for 8 uniformly sampled frames of each video in the ActivityNet train split, we generate bounding box annotations for all objects belonging to COCO dataset categories and a list of COCO dataset categories not present in those 8 frames. This data is sufficient to implement our IFT objectives on the ActivityNet video dataset with only the videos from the dataset. Our promising results (see \cref{tbl:vid_more}) for video-domain IFT using only pseudo-data highlight the data scalability of our proposed framework.

\section{Video Architecture \& Training}
\label{supp:video}
As discussed in \cref{subsec:video} (main paper), we introduce two video-domain variants of our framework, \modelname-Vid-B and \modelname-Vid-B+. We first detail the architecture common to both variants, followed by specific training procedures. 

The overall architecture remains consistent to what is presented in \cref{fig:llava}. The visual encoder processes $n_f$ frames independently as images to produce $n_f \times 256$ visual tokens per video (where 256 is tokens generated per image). The spatio-temporal pooling strategy from \cite{Maaz2023VideoChatGPT} is utilized to obtain a set of $256 + n_f$ visual tokens per video. In detail, the visual tokens are average pooled across the temporal dimension to obtain $256$ spatial tokens and across the spatial dimensions to obtain $n_f$ temporal tokens. These are concatenated to obtain the $256 + n_f$ visual tokens per video. The adaptor layer and LLM remain unchanged - this is straightforward since both these layers perform set-to-set operations independent of input sequence length. 

The \modelname-B-Vid+ variant combines our video level IFT objectives with the training setup from \cite{Maaz2023VideoChatGPT}. Given early experiments suggesting insufficiency of fine-tuning only the adapter layer for our IFT objectives, we fine-tune both the LLM and the adaptor layer. We also sample only 8 uniformly spaced frames per video (for compute reasons). 
The three IFT objectives are modified to suit video domain operation. Given the lack of explicit temporal modelling in our visual backbone and the limited spatio-temporal awareness even within the LLM, we focus on static objects in videos to construct IFT targets. For LocPred and RevLoc, we first filter out objects to select those present only in one of the eight frames or relatively static ones (bounding-box center (x,y) is within a 5 pixel range from their average if present in multiple frames). Then, we obtain the average bounding-box for that object across the frames. These static bounding boxes and negative categories (from the dataset) are used to construct the IFT targets in the same manner as we do for images.

\section{Spatial Reasoning Toy Experiment}
\label{supp:toy}
We present additional details of the toy experiment introduced in \cref{subsec:spatial}. We describe the dataset used for evaluation, templates for prompting, and evaluation metric calculation. We also repeat our results from \cref{tbl:spatial_icl} (main paper) for the left vs right variant here in \cref{app:spatial_icl}. 

We first construct an evaluation dataset, tagged \textit{COCO-Spatial-27K} containing 26,716 image-question pairs. We build this off the COCO dataset \cite{lin2014microsoft} train split through a fully-automated process, utilizing the ground-truth object bounding-box annotations. We first filter out images based on three constraints - this eliminates a large portion of images; hence we elect to use the train split to obtain a considerable quantity of samples after filtering. We first select images containing distinct category object triplets (only one instance occurrence of each object category). For example, an image would contain categories person, dog, and table but only one of each. The second constraint ensures that each object is entirely to the left or right half of the image. This is based on object center not being in the central 20\% region. The third constraint is that at least two objects are on opposite sides (i.e. left and right half of image). This provides at least two opposite side object pairs. The ground-truth bounding box annotations enable easy automation of this filtering procedure.

We next discuss our templates for prompting. For two objects on opposite sides tagged \texttt{obj\_1} and \texttt{obj\_2}, we use the prompt \texttt{Which side of obj\_1 is obj\_2 located?} and query the model for a response. This is for the direct VQA setting. In the case of in-context learning (ICL) VQA setting, we preprend two examples to the prompt: \texttt{Q: Which side of obj\_1 is obj\_2 located? A: The obj\_1 is located to the left of obj\_2. Q: Which side of obj\_2 is obj\_1 located? A: The obj\_2 is located to the right of obj\_1. Q: Which side of obj\_3 is obj\_1 located?}. In this case, \texttt{obj\_3} is the third object, and their ordering is selected such that \texttt{obj\_1} is on one side, and \texttt{obj\_2}, \texttt{obj\_3} are on the opposite side. 

Building off standard VQA protocol in \cite{Hudson2019GQAAN,Maaz2023VideoChatGPT}, we simply query is the terms \texttt{left} or \texttt{right} are present in the generated outputs, and rate it a success is the target term is present in the generated response. 
We also visualize some examples for this task in \cref{fig:vis_spatial}.

\begin{table}[t]
\centering
\small
\def\arraystretch{1.0}  
\setlength\tabcolsep{0.65em}  
\begin{tabular}{l|c|c|c|c}
\toprule
Method      & ICL       & Acc (All)& Acc (Left)& Acc (Right)\\ \midrule
BLIP-2 \cite{li2023blip}     &  \xmark   & 45.5     & 86.1      & 4.74       \\
LLaVA  \cite{liu2023visual}     &  \xmark   & 55.1     & 84.5      & 36.5       \\ \rowcolor{Gray}
Ours        &  \xmark   & 69.5     & 79.7      & 59.2       \\ \midrule
BLIP-2 \cite{li2023blip}     &  \cmark   & 14.7     & 17.8      & 11.6       \\ 
LLaVA \cite{liu2023visual}      &  \cmark   & 55.1     & 84.7      & 36.4       \\ \rowcolor{Gray}
Ours        &  \cmark   & 76.5     & 90.4      & 61.5       \\ \bottomrule
\end{tabular}
\vspace{-0.5em}
\caption{\textbf{Spatial Reasoning}: We repeat our results for left vs right objects here.}
\label{app:spatial_icl}
\vspace{-1.0em}
\end{table}

\section{LLaVA Dataset Analysis}
\label{supp:llava_dataset}

Our results in \cref{app:spatial_icl} indicate unusual disparity in left vs right accuracy numbers, especially in LLaVA \cite{liu2023visual}. We analyse the training dataset used in this LLaVA baseline to better understand these disparities.

The LLaVA model \cite{liu2023visual} is instruction fine-tuned on a human conversation style dataset (LLaVA-Instruct-80K). This dataset contains 80,000 image-conversation pairs leading to 221,333 question-answer (QA) pairs across all images (multiple QA for single image). We analyse the presence of keywords related to \texttt{left} and \texttt{right} concepts that are probed in our spatial-reasoning toy experiment (\cref{subsec:spatial}).  

We first analyse the exact presence of the words \textit{left} and \textit{right} in the corpus (noting this maybe in different context, e.g. who has the right of way?). Of the 80,000 image-conversation pairs, left and right are present in 1619 (2.02\%) and 5001 (6.25\%) cases respectively. We provide further statistics of the dataset in \cref{tbl:supp_llava} indicating some presence of conversation style training samples encompassing left \& right concepts. A large count of the keyword \texttt{right} occurs in contexts with different meanings while \texttt{left} mostly occurs in its spatial context. We hypothesize that this may be the reason for predicting \texttt{left} more often when models are queried with a spatial reasoning related question (i.e. keyword \texttt{left} occurs more frequently with \textit{spatial related words} in training corpus). 

\begin{table}[h]
\vspace{-0.5em}
\centering
\small
\def\arraystretch{1.1}  
\setlength\tabcolsep{1.0em}  
\begin{tabular}{l|c|c} 
\toprule
Template             & Left (\%)  & Right (\%)  \\ \midrule
``the \{\} ''        & 171 (0.21) & 1314 (1.54) \\
``\{\} side''        & 75 (0.093) & 110 (0.14)  \\
``to the \{\}''      & 80 (0.10)  & 93 (0.12)   \\ \bottomrule
\end{tabular}
\vspace{-0.8em}
\caption{We count occurrences of various textual phrases related to left \& right concepts in the LLaVA-Instruct-80K dataset.}
\label{tbl:supp_llava}
\vspace{-0.5em}
\end{table}

\noindent
Therein, we attribute these observed disparities for left vs right accuracy numbers to these artifacts present in datasets used for training underlying LLMs.

\section{Limitations \& Broader Impact}
Our video variant achieves strong performance on VQA tasks but fails to understand temporal locations. In fact, direction use of temporal locations paired with spatial locations results in training collapse for our framework. Extension of our instruction fine-tuning objectives to suitably utilize time coordinates is left as a future direction. 
In terms of broader impact, while our model uses generic vision and language model architectures, we note that our training data from public datasets may contain biases which should be taken into account when deploying models trained using our framework.

\section{Qualitative Evaluation}
In this section, we present visual examples showcasing various aspects of our frameworks capabilities. We broadly consider the three distinct settings of spatial reasoning, region description, and generated locations. Note that in all visualizations we blur human faces to make them unidentifiable for privacy reasoings. 

\begin{figure*}
    \centering
    \includegraphics[width=\linewidth]{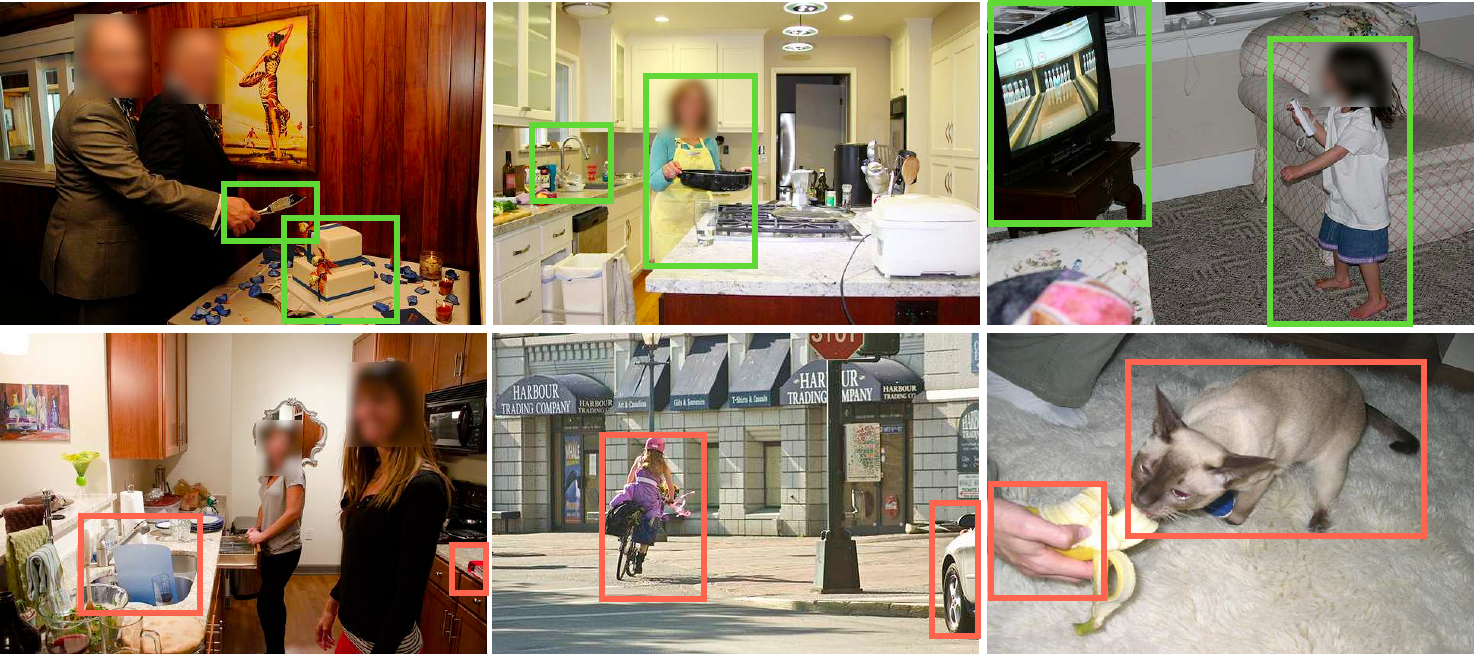}
    \vspace{-2.0em}
    \caption{\small 
    Visualizing Spatial Reasoning: We illustrate example images on which we perform our toy experiment for spatial reasoning (\cref{supp:toy}). Success cases on top row (green) and failure cases on bottom row (red).}
    \label{fig:vis_spatial}
    \vspace{-0.5em}
\end{figure*}

\begin{figure*}
    \centering
    \includegraphics[width=\linewidth]{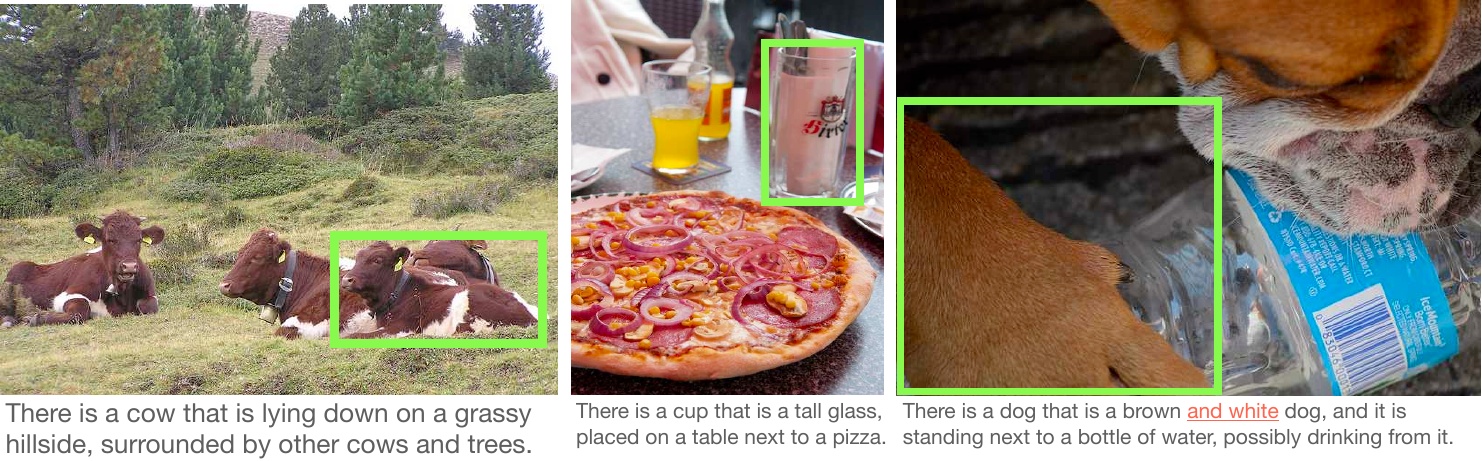}
    \vspace{-1.7em}
    \caption{\small 
    Visualizing Region Description: Our framework possesses the unique ability of generating representative descriptions for a selected region of an image, input to the model in terms of textual coordinates. We illustrate 3 example images with a bounding box (green) denoting the queried region. The responses generated by our model are underneath each image, with invalid outputs \todo{\underline{highlighted red}}.  
    }
    \label{fig:vis_rd}
    \vspace{-0.7em}
\end{figure*}

\begin{figure*}
    \centering
    \includegraphics[width=\linewidth]{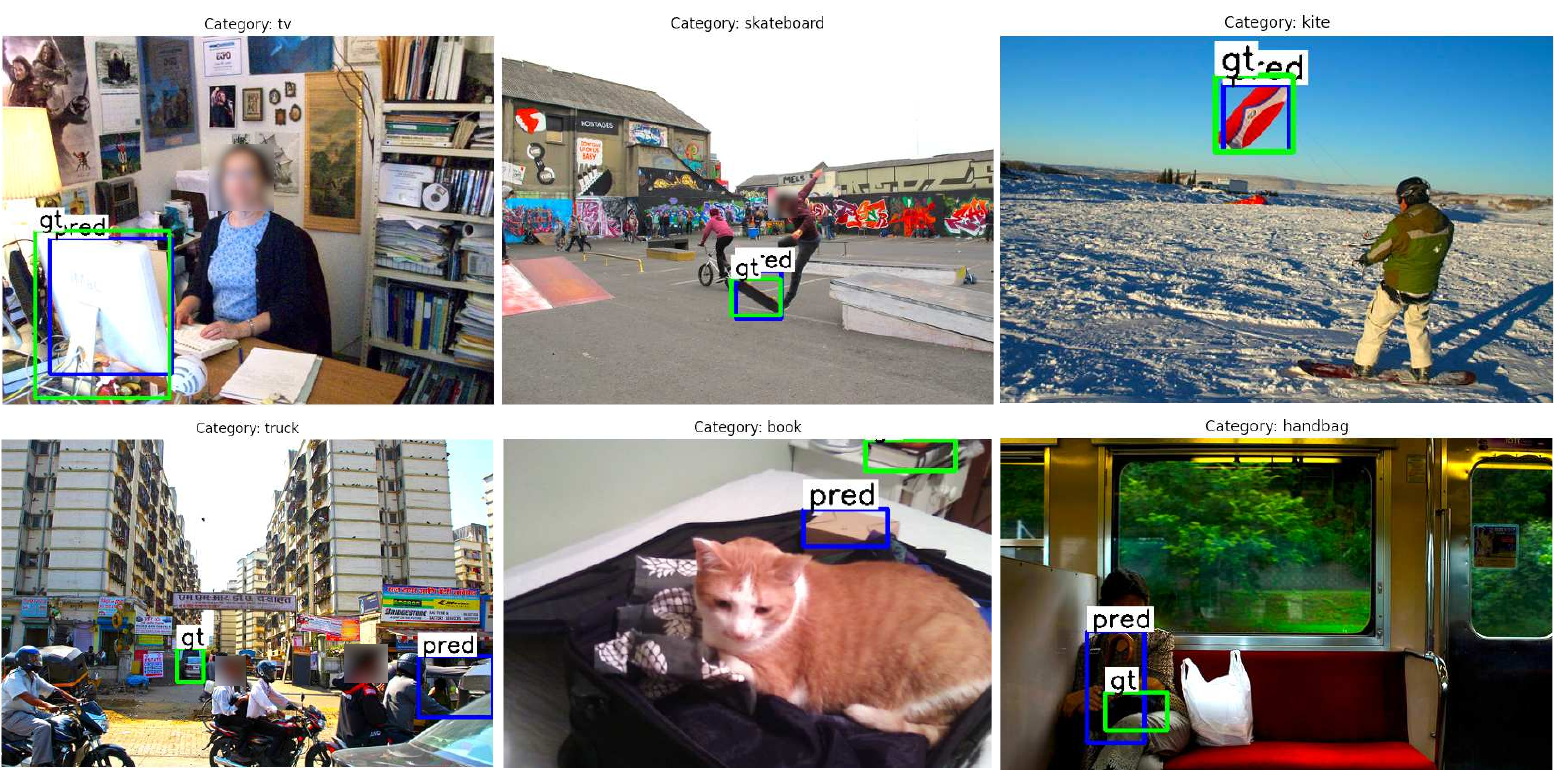}
    \vspace{-2.0em}
    \caption{Visualization of LocPred Objective: We illustrate the bounding box locations generated by our framework (blue) when queried with a category label (top of each image) and compare with the ground-truth bounding boxes (green). Success cases on top and failure cases on bottom.}
    \label{fig:vis_loc}
    \vspace{-0.5em}
\end{figure*}

\noindent
\textbf{Spatial Reasoning:} We illustrate examples from our COCO-Spatial-27K dataset highlighting both success cases and failures of our framework. These qualitative results are presented in \cref{fig:vis_spatial}. In each case, let as tag the two objects within bounding boxes as \texttt{obj1} and \texttt{obj2}. Following \cref{supp:toy}, we prompt our framework with each image and \texttt{Which side of obj1 is obj2?} and match the response with the ground-truth answer. Correct matches (success cases) on presented on the top row (green) and incorrect matches (failure cases) on bottom row (red). The correct matches indicate the spatial reasoning abilities of our framework across a wide range of image types, including cluttered scenes. The failure cases possibly indicate difficulty at handling truncated / occluded objects. 

\noindent
\textbf{Region Description:}
We next illustrate the region description abilities of our model (see \cref{subsec:revloc} for details) in \cref{fig:vis_rd}. We query our framework with a set of bounding box coordinate such as \texttt{Describe the object located at [22, 114, 86, 154]?} (prompt details in \cref{supp:prompt}) paired with each image. We illustrate the object coordinates as a bounding box (green) in each image. The response of the model presented underneath each image. We highlight invalid responses in \todo{\underline{red}}. These qualitative evaluations indicate the ability of our model to not only detect the object present in the queried region, but also describe it in terms of its surrounding: an ability unique to our model in contrast to traditional object classifiers or detectors. At the same time, the generated responses display limitations in terms of object characteristic hallucination and minimal spatial relation (e.g. to the left / right of) based description. 

\noindent
\textbf{Generated Locations:}
In our experiments, the tasks of object hallucination and region description directly evaluate the learning resulting from IFT objectives NegPred and RevLoc respectively. In this section, we present some qualitative evaluation to understand the learning resulting from the LocPred objective. These results are visualized in \cref{fig:vis_loc}. First, these images present samples from the validation split of COCO modified in a similar manner (i.e. filtering explain in \cref{sec:exp}) to our training set for LocPred objective. Each image contains one instance of a particular category. The category is labelled on top of each image, and the ground-truth annotation for the object is in green while the prediction by our framework is in blue. We illustrate the success cases of our model in the top row and failure cases in the bottom row. The success cases indicate strong localization skills across diverse scene involving objects of variable sizes. The failure cases denote difficulty in handling crowded / cluttered scenes and truncated / occluded objects. We also note that direct comparison to classical object detectors is unfair given the down-sampled images (i.e. $224 \times 224$ or $336 \time 336$ sized) used by our framework (object detectors use higher resolution images). 